\documentclass[11pt,a4paper]{article}
\usepackage[hyperref]{acl2017}
\usepackage{times}
\usepackage{latexsym}
\usepackage{graphicx}
\usepackage{url}
\usepackage{booktabs}
\usepackage{fancyvrb}

\aclfinalcopy 

\title{The Meaning Factory at SemEval-2017 Task 9: Producing AMRs with Neural Semantic Parsing}

\author{Rik van Noord \\
  CLCG\\
  University of Groningen \\
  {\tt r.i.k.van.noord@rug.nl} \\\And
  Johan Bos \\
  CLCG\\
  University of Groningen \\
  {\tt johan.bos@rug.nl} \\}

\date{}

\begin{document}
\maketitle
\begin{abstract}
We evaluate a semantic parser based on a character-based sequence-to-sequence model in the context of the SemEval-2017 shared task on semantic parsing for AMRs. With data augmentation, super characters, and POS-tagging we gain major improvements in performance compared to a baseline character-level model. Although we improve on previous character-based neural semantic parsing models, the overall accuracy is still lower than a state-of-the-art AMR parser. An ensemble combining our neural semantic parser with an existing, traditional parser, yields a small gain in performance.
\end{abstract}

\section{Introduction}

Traditional open-domain semantic parsers often use statistical syntactic parsers to derive syntactic structure on which to build a meaning representation. Recently there have been interesting attempts to view semantic parsing as a translation task, mapping a source language (here: English) to a target language (a logical form of some kind). \newcite{dong2016language} used sequence-to-sequence and sequence-to-tree neural translation models to produce logical forms from sentences, while \newcite{riga:16} and \newcite{peng:17} used a similar method to produce AMRs. From a purely engineering point of view, these are interesting attempts as complex models of the semantic parsing process can be avoided. Yet little is known about the performance and fine-tuning of such parsers, and whether they can reach performance of traditional semantic parsers, or whether they could contribute to performance in an ensemble setting.

In the context of SemEval-2017 Task 9 we aim to shed more light on these questions. In particular we participated in 
Subtask 1, Parsing Biomedical Data, and work with parallel English-AMR training data comprising extracts of
scientific articles about cancer pathway discovery. 

  



More specifically, our objectives are (1) try to reproduce the results of \newcite{riga:16}, who used
 character-level models for neural semantic parsing; (2) improve on their results by employing several novel techniques; and (3) combine the resulting neural semantic parser with an existing off-the-shelf AMR parser to reach state-of-the-art results.

\section{Neural Semantic Parsing}

\subsection{Datasets}

Our training set consists of the second LDC AMR release (LDC2016E25) containing 39,620 AMRs, as well as the training set of the bio AMR corpus that contains 5,452 AMRs. As development and test set we use the designated development and test partition of the bio AMR corpus, both containing 500 AMRs. HTML-tags are removed from the sentences.

\subsection{Basic Translation Model}

We use a seq2seq neural translation model to \emph{translate} English sentences into AMRs. This is a bi-LSTM model with added attention mechanism, as described in \newcite{bahdanau2014neural}. Similar to \newcite{riga:16}, but contrasting with \newcite{peng:17}, we train the model only on character-level input. Model specifics are shown in Table \ref{tab:model}.

\begin{table}[ht]
\centering
\begin{tabular}{l|l}
\toprule
\textbf{Parameter}    & \textbf{Value}\\
\midrule
Layers        & 1         \\
Nodes         & 400       \\
Buckets       & (510,510) \\
Epochs        & 25--35     \\
Vocabulary    & 150--200   \\
Learning rate & 0.5  \\ 
Decay factor & 0.99  \\ 
Gradient norm & 5  \\ 
\bottomrule
\end{tabular}
\caption{Model specifics for the seq2seq model.\label{tab:model}}
\end{table}

In a preprocessing step, we remove all variables from the AMR and duplicate co-referring nodes. An example of this is shown in Figure \ref{fig:amrseq}. The variables and co-referring nodes are restored after testing, using the restoring script from \mbox{\newcite{riga:16}}.\footnote{https://github.com/didzis/tensorflowAMR}
Wikipedia links are also removed from the training set, but get restored in a separate Wikification post-processing step.


\begin{figure}[ht]
{\small
\begin{minipage}{\textwidth}
  \begin{Verbatim}[commandchars=\\\{\}]
(require-01
   :ARG0 (induce-01
      :ARG1 (cell)
      :ARG2 (migrate-01
         \textbf{:ARG0 cell))}
   :ARG1 (bind-01
      :ARG1 (protein 
         :name (name :op1 "Crk"))
      :ARG2 (protein 	
         :name (name :op1 "CAS"))))
\end{Verbatim}
  \end{minipage}
  \caption{\label{fig:amrseq}\emph{``Crk binding to CAS is required for the induction of cell migration''} - seq2seq tree representation.}
}
\end{figure} 

\subsection{Improvements}

In this section we describe the methods used to improve the neural semantic parser.

\paragraph{Augmentation}

AMRs, as introduced by \newcite{Banarescu13abstractmeaning}, are rooted, directed, labeled graphs, in which the different nodes and triples are unordered by definition. However, in our tree representation of AMRs (see Figure \ref{fig:amrseq}), there is an order of branches. This means that we are able to permute this order into a more intuitive representation of the sentence, by matching the word order using the AMR-sentence alignments. An example of this method is shown in Figure \ref{fig:amrseqmatch}.
  
\begin{figure}[ht]
{\small
\begin{minipage}{\textwidth}
  \begin{Verbatim}[commandchars=\\\{\}]
(require-01
   \textbf{:ARG1 (bind-01}
      :ARG1 (protein 
         :name (name :op1 "Crk"))
      :ARG2 (protein 	
         :name (name :op1 "CAS"))))
   \textbf{:ARG0 (induce-01}
      :ARG1 (cell)
      :ARG2 (migrate-01
         :ARG0 cell))
\end{Verbatim}
  \end{minipage}
  \caption{\label{fig:amrseqmatch}\emph{``Crk binding to CAS is required for the induction of cell migration''} - seq2seq representation that best matches the word order.}
} 
\end{figure}

This approach can also be used to augment the training data, since we are now able generate ``new'' AMR-sentence pairs that can be added to our training set. However, due to the exponential growth, there are often more than 1,000 different AMR permutations for long sentences. We ran multiple experiments to find the best way to use this oversupply of data. Ultimately, we found that it is most beneficial to ``double'' the training data by adding the best matching AMR (based on word order) to the existing data set.

\paragraph{Super characters}

We do not necessarily have to restrict ourselves to using only individual characters as input. For example, the AMR relations (e.g. \texttt{:ARG0}, \texttt{:domain}, \texttt{:mod}) can be seen as single entities instead of a collection of characters. This decreases the input length of the AMRs in feature space, but increases the total vocabulary. We refer to these entities as \emph{super characters}. This way, we essentially create a model that is a combination of character and word level input. 

\paragraph{POS-tagging}

Character-level models might still be able to benefit from syntactic information, even when this is added directly to the input structure. Especially POS-tags can easily be added as features to the input data, while also providing valuable information. For example, proper nouns in a sentence often occur with the \texttt{:name}
relation in the corresponding AMR, while adjectives correlate with the \texttt{:mod} relation. We append the corresponding POS-tag to each word in the sentence (using the C\&C POS-tagger by \newcite{clark2003bootstrapping}), creating a new super character for each unique tag.

\paragraph{Post-processing}

In a post-processing step, first the variables and co-referring nodes are restored. We try to fix invalidly produced AMRs by applying a few simple heuristics, such as inserting special characters (e.g. parentheses, quotes) or removing unfinished edges. If the AMR is still invalid, we output a smart default AMR.\footnote{This was not necessary for the evaluation data.} 

We also remove all double nodes, i.e., relation-concept pairs that occur more than once in a branch of the AMR. This form of duplicate output is a common problem with deep learning models, since the model does not keep track of what it has already output. We refer to this method as \emph{pruning}.

\paragraph{Wikification}

Our Wikification method is based on \newcite{bjerva2016meaning}, using Spotlight \cite{spotlight:2013}. They initially removed wiki links from the input and then tried to restore them in the output by simply adding a wiki link to the AMR if it matches with the name in a \texttt{:name} relation. Even though this approach worked well for the LDC data, it did not work for the biomedical data.

This is mainly due to the fact that \texttt{:name} nodes are not consistently annotated with a wiki link in the gold biomedical data. 138 unique names that had a corresponding wiki link at least once in the gold data did not have this wiki link 100\% of the time. For example \emph{DNA} occurred 86 times as a \texttt{:name} in the gold data and was only annotated with a wiki link in 69 cases, while \emph{ERK} occurred 228 times with only 3 annotated wiki links.
%
%
For this reason we opt for a safe Wikification approach: we only add wiki links to names that were annotated with the same wiki link more than 50\% of the time in the gold data. Following our previous example, this means that \emph{DNA} does still get a wiki link, but \emph{ERK} does not.

\subsection{CAMR ensemble}

As we know that our neural semantic parser is unlikely to outperform a state-of-the-art AMR parser, but is likely to complement it, our strategy is to use an ensemble-based approach. The ensemble comprises the off-the-shelf parser CAMR \cite{CAMR:15}
and our neural semantic parser. 
The implementation of this ensemble is similar to \newcite{riga:16}, choosing the AMR that obtains the highest pairwise Smatch \cite{smatch:2013} score when compared to the other AMRs generated for a sentence. This method is designed to ultimately choose the AMR with the most prevalent relations and concepts.

We train CAMR models based on the biomedical data only, the LDC data only and the combination of both data sets. Since CAMR is nondeterministic, we can also train multiple models on the same data set. Ultimately, the best ensemble on the test data consisted of three bio-only models, two bio + LDC models and one LDC-only model. This ensemble was used to parse the evaluation set.

\section{Results and Discussion}

\subsection{Results on Test Set}

Table~\ref{tab:restest} shows the results of all improvement methods tested in isolation on the test set of the biomedical data. Augmenting the data only helps very slightly, while the super characters are responsible for the largest increase in performance. This shows that we do not necessarily have to use only character or word level input in our models, but that a combination of the two might be optimal. The best result was obtained by combining the different methods. This model was then used to parse the evaluation data. Table \ref{tab:rescamr} shows the results of retraining CAMR on different data sets, as well as an ensemble of those models. Adding our seq2seq model to the ensemble only yielded a very small gain in performance.

\begin{table}[hbtp]
\centering
\begin{tabular}{l|l|l}
\toprule                                                                    \textbf{Feature} & \textbf{F-score} & \textbf{Increase} \\ \midrule
Baseline &          0.422        &                   \\
Pruning    &        0.425          &  0.7\%                 \\
Wikification  &     0.423             & 0.2\%     \\
Augmentation              &  0.424                & 0.5\%     \\
Super characters          &  0.481                & 14.0\%     \\
POS-tagging               &  0.436                & 3.3\% \\ 
\midrule
All combined                    &  0.504                &  19.4\%
\\ \bottomrule
\end{tabular}
\caption{\label{tab:restest}Results of the different seq2seq models on the test set of the biomedical data.}
\end{table}

\begin{table}[hbtp]
\centering
\begin{tabular}{l|l}
\toprule
                            & \textbf{F-score} \\
\midrule
CAMR retrained on LDC       & 0.399   \\
CAMR retrained on bio       & 0.585   \\
CAMR retrained on LDC + bio & 0.582   \\
\midrule
Ensemble CAMR               & 0.588   \\
Ensemble CAMR + seq2seq  & 0.589  \\
\bottomrule
\end{tabular}
\caption{\label{tab:rescamr}Results of retraining CAMR and results of best ensemble models, tested on the biomedical test data.}
\end{table}

\subsection{Official Results}

In Table~\ref{tab:reseval} we see the detailed results of the best seq2seq model and best ensemble on the evaluation data, using the scripts from \newcite{damonte:17}.\footnote{Unofficial score for seq2seq negation; due to a mistake, all \texttt{:polarity} nodes were removed in the official submission. This had no influence on the final F-score.} While CAMR has similar scores on the test data, the score of the seq2seq model decreases by 0.04. It is interesting to note that seq2seq scores equally well without word sense disambiguation, while there is no separate module that handles this. 

\begin{table}[h]
\centering
\begin{tabular}{l|r|r}
\toprule
 \textbf{Setting}              & \textbf{seq2seq} & \textbf{Ensemble} \\ \midrule
\textbf{Smatch}         & \textbf{0.460}                & \textbf{0.576}      \\
Unlabeled      & 0.504                 & 0.623              \\ 
No WSD        & 0.463                & 0.579              \\
Named Entities & 0.512                & 0.576              \\
Wikification   & 0.458                & 0.396              \\
Negation       & 0.141                & 0.244              \\
Concepts       & 0.630                & 0.759              \\
Reentrancies   & 0.290                & 0.352              \\
SRL            & 0.427                & 0.543             \\
\bottomrule
\end{tabular}
\caption{\label{tab:reseval}Official results on the evaluation set for both the ensemble and the seq-to-seq neural semantic parser.}
\end{table}

\begin{figure}[h!]
  \centering
  \includegraphics[scale=0.41]{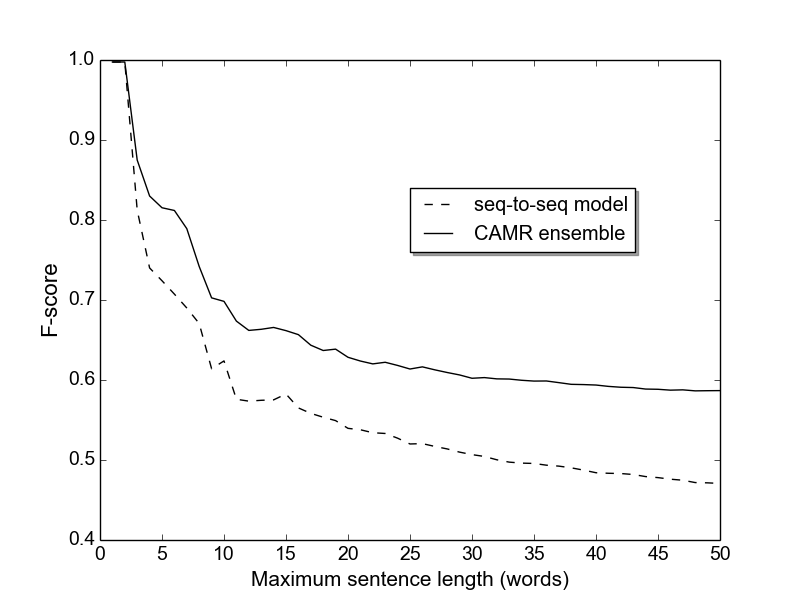}
  \caption{\label{fig:senlength}Comparison of CAMR and our seq-to-seq model for different sentence lengths.}
\end{figure}

\subsection{Comparison with CAMR}

Although CAMR outperformed our neural semantic parser by a large margin, the seq-to-seq model did produce a better AMR for 108 out of the 500 evaluation AMRs, based on Smatch score. 
If the CAMR + seq2seq ensemble was somehow able to always choose the best AMR, it obtains an F-score of $0.601$, an increase of 2.2\% instead of the current 0.2\%. This suggests 
that the current method of combining neural semantic parsers with existing parsers is far from optimal, but that the neural methods do provide complementary information. A different way to incorporate this information would be to pick the most suitable parser based on the input sentence. A classifier that exploits the characteristics of the sentence could be trained to assign a parser to each individual (to be parsed) sentence.

Figure \ref{fig:senlength} shows the performance of the neural semantic parser and the CAMR ensemble per maximum sentence length. We see that seq-to-seq can keep up with CAMR for very short sentences, but is clearly outperformed on longer sentences. As the sentences get longer, the difference in performance gets bigger, but not much.

\section{Conclusion and Future Work}

We were able to reproduce the results of the character-level models for neural semantic parsing as proposed by \newcite{riga:16}. Moreover, we showed improvement on their basic setting by using data-augmentation, part-of-speech as additional input, and using super characters.  The latter setting showed that a combination of character and word level input might be optimal for neural semantic parsers.
Despite these enhancements, the resulting AMR parser is still outperformed by more traditional, off-the-shelf AMR parsers.
Adding our neural semantic parser to an ensemble including CAMR \cite{CAMR:15}, a dependency-based parser, yielded no noteworthy improvements on the overall performance.

Do these results indicate that neural semantic parsers will never be competitive with more traditional statistical parsers? We don't think so. We have the feeling that we have just scratched the surface of possibilities that neural semantic parsing can offer us, and how they possibly can complement parsers using different strategies. In future work we will explore these.

\bigskip
\subsection*{Acknowledgements} 

We thank the two anonymous reviewers for their comments. We would also like to thank the Center for Information Technology of the University of Groningen for their support
and for providing access to the Peregrine high performance computing cluster. We also used a
Tesla K40 GPU, which was kindly donated by the NVIDIA Corporation. This work was funded by the NWO-VICI grant ``Lost in Translation -- Found in Meaning'' (288-89-003). 

\bibliography{semeval2017}
\bibliographystyle{acl_natbib}

\end{document}